\pdfoutput=1

\documentclass[11pt]{article}

\usepackage{acl}

\usepackage{times}
\usepackage{latexsym}
\usepackage{amsthm}
\usepackage{booktabs}
\usepackage{algorithm}
\usepackage{algorithmic}
\usepackage{multirow}
\usepackage{amsmath, amssymb}
\usepackage{wasysym}
\usepackage{bbm}
\usepackage{mathtools}
\usepackage{tablefootnote}
\usepackage{subfigure}
\usepackage{pifont}
\usepackage{enumitem}

\usepackage[T1]{fontenc}

\usepackage[utf8]{inputenc}

\usepackage{microtype}

\title{Revisiting Event Argument Extraction: Can EAE Models Learn Better When Being Aware of Event Co-occurrences?}

\makeatletter
\def\thanks#1{\protected@xdef\@thanks{\@thanks
        \protect\footnotetext{#1}}}
\makeatother

\author{Yuxin He$^1$ \and Jingyue Hu$^1$ \and Buzhou Tang\thanks{\textsuperscript{$\dagger$}Corresponding Author.}$^{1, 2, \dagger}$ \\
        \textsuperscript{1}{Department of Computer Science, Harbin Institute of Technology, Shenzhen, China}\\
        \textsuperscript{2}{Peng Cheng Laboratory, Shenzhen, China}\\
        \texttt{21S051047@stu.hit.edu.cn}\\
        \texttt{tangbuzhou@gmail.com}
}

\begin{document}
\maketitle
\begin{abstract}
Event co-occurrences have been proved effective for event extraction (EE) in previous studies, but have not been considered for event argument extraction (EAE) recently. In this paper, we try to fill this gap between EE research and EAE research, by highlighting the question that \emph{``Can EAE models learn better when being aware of event co-occurrences?''}. To answer this question, we reformulate EAE as a problem of table generation and extend a SOTA prompt-based EAE model into a non-autoregressive generation framework, called TabEAE, which is able to extract the arguments of multiple events in parallel. Under this framework, we experiment with 3 different training-inference schemes on 4 datasets (ACE05, RAMS, WikiEvents and MLEE) and discover that via training the model to extract all events in parallel, it can better distinguish the semantic boundary of each event and its ability to extract single event gets substantially improved. Experimental results show that our method achieves new state-of-the-art performance on the 4 datasets. Our code is avilable at \url{https://github.com/Stardust-hyx/TabEAE}.
\end{abstract}

\section{Introduction}

Event argument extraction (EAE) is an essential subtask of event extraction (EE). Given an input text and trigger(s) of target event(s), the EAE task aims to extract all argument(s) of each target event.  Recently, substantial progress has been reported on EAE, thanks to the success of pre-trained language models (PLMs).

\begin{figure}[t]
\centering
\includegraphics[width=0.9\linewidth]{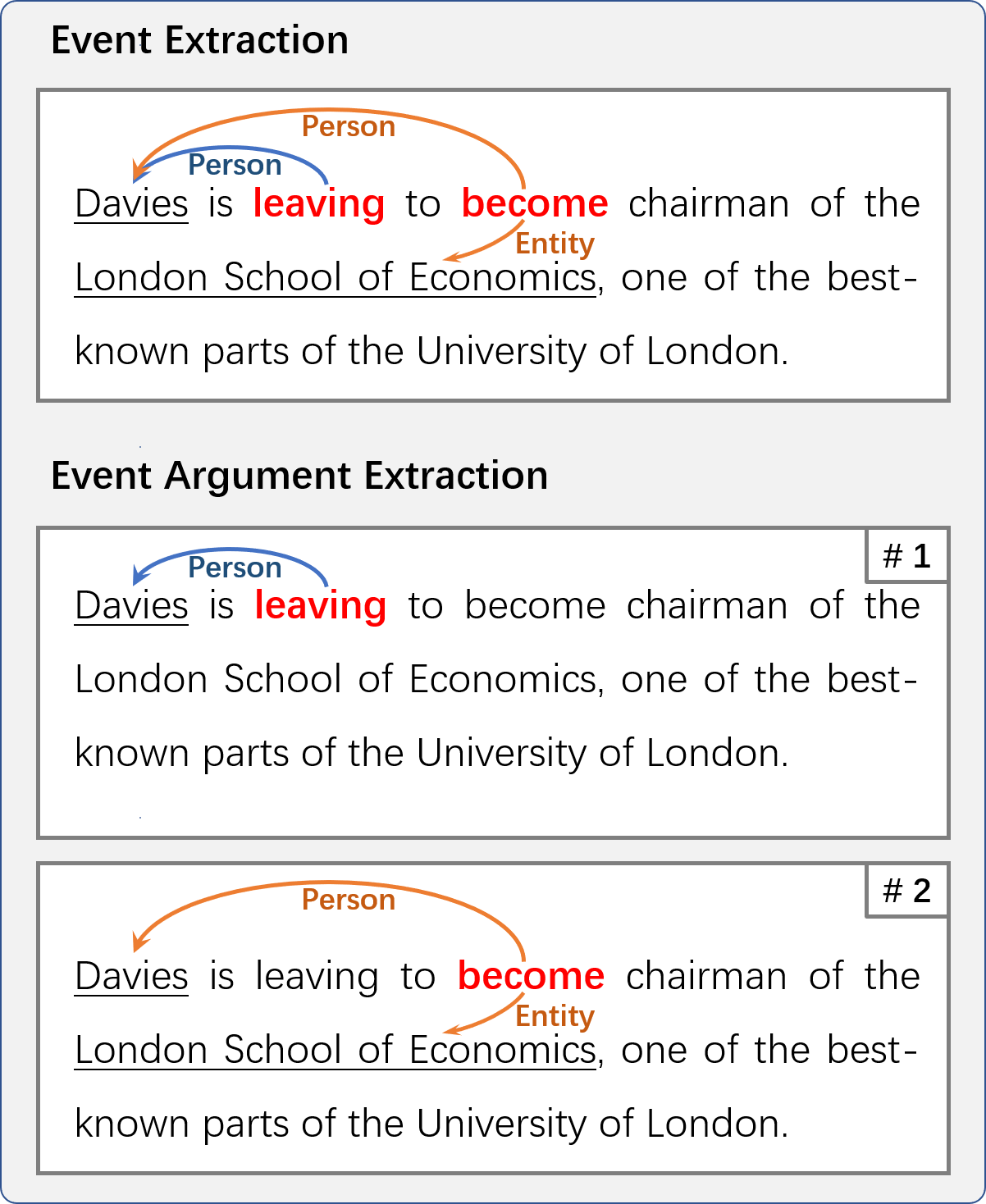}

\caption{An illustration of EE and EAE. The triggers are in red and the arguments are underlined. EE models aim at extracting all events concurrently, whereas mainstream EAE models are trained to extract the arguments for one event trigger at a time.}
\label{intro}
\vspace{-0.5em}
\end{figure}

Previous studies on EE commonly take event co-occurrences into account. However, recent works on EAE \citep{ebner-etal-2020-multi, zhang-etal-2020-two, xu-etal-2022-two, du-cardie-2020-event, wei-etal-2021-trigger, liu-etal-2021-machine, li-etal-2021-document, du-etal-2021-template, lu-etal-2021-text2event, ma-etal-2022-prompt} only consider one event at a time and ignore event co-occurrences (as illustrated in Figure \ref{intro}). In fact, event co-occurrences always exisit in text and they are useful in revealing event correlation and contrasting the semantic structures of different events. For the instance in Figure \ref{intro}, there exist two events in the same context. The two events are triggered by ``leaving'', ``become'' respectively, and share the same subject ``Davies''. It is clear that there exists a strong causal correlation between the two events. However, mainstream works on EAE split the instance into two samples, which conceals this correlation.

In this paper, we try to resolve this divergence between EE research and EAE research, by highlighting the question that \emph{``Can EAE models learn better when being aware of event co-occurrences?''}. To address this question, we reformulate EAE as a problem of table generation and extend the SOTA prompt-based EAE model, PAIE \cite{ma-etal-2022-prompt}, into a non-autoregressive generation framework to extract the arguments of multiple events concurrently. Our framework, called TabEAE, inherits the encoding, prompt construction and span selection modules from PAIE, but employs a novel non-autoregressive decoder for table generation.



Under this framework, we explore three kinds of training-inference schemes: (1) \emph{Single-Single}, training model to extract single event at a time and infer in the same way; (2) \emph{Multi-Multi}, training model to extract all events in parallel and infer in the same way; (3) \emph{Multi-Single}, training model to extract all events in parallel and let the model extract single event at a time during inference.
According to our experiments, the Multi-Single scheme works the best on 3 benchmarks (ACE, RAMS and WikiEvents) and the Multi-Multi scheme works the best on the MLEE benchmark, where the phenomenon of nested events extensively exists. Besides, in-depth analysis reveals that via training TabEAE to extract all events in parallel, it can better capture the semantic boundary of each event and its ability to extract single event at a time gets substantially improved.

To sum up, our contributions include:
\begin{itemize}[leftmargin=*]
    \setlength{\itemsep}{2pt}
    \setlength{\parsep}{2pt}
    \setlength{\parskip}{2pt}
    \item We observe the divergence between EE research and EAE research in terms of the phenomenon of event co-occurrence. To resolve this divergence, we extend the SOTA prompt-based EAE model PAIE into a text-to-table framework, TabEAE, which is able to extract the arguments of multiple events concurrently.
    \item Under the TabEAE framework, we explore three training-inference schemes, i.e. Single-Single, Multi-Multi, Multi-Single, and verify the significance of event co-occurrence for EAE. 
    \item The proposed method outperforms SOTA EAE methods by 1.1, 0.4, 0.7 and 2.7 in Arg-C F1 respectively on the 4 benchmarks ACE05, RAMS, WikiEvents and MLEE.
\end{itemize}

\begin{figure*}[ht]
	\centering
	\includegraphics[width=\textwidth]{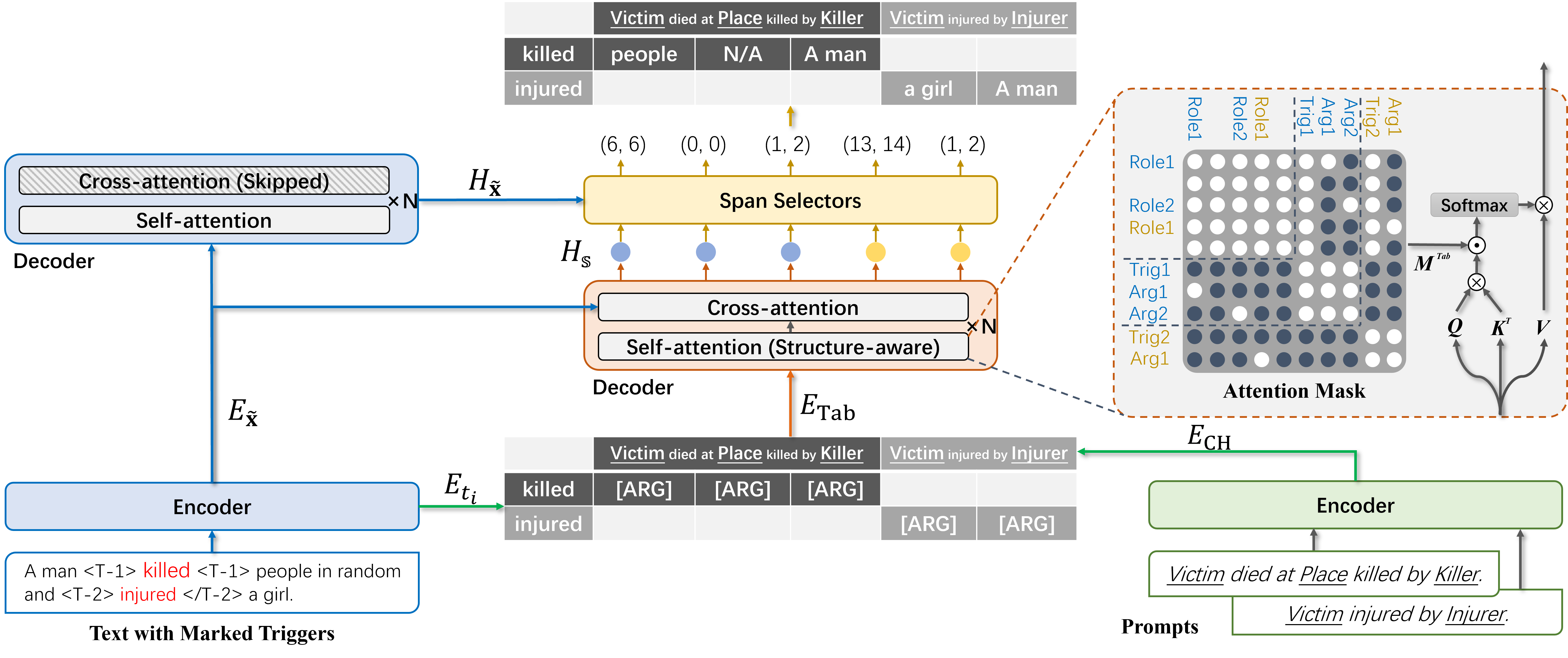}
	\caption{An overview of TabEAE, our non-autoregressive table generation framework. It includes four components: (1) Trigger-aware context encoding (in blue); (2) Slotted table construction (in green); (3) Non-autoregressive table decoding (in orange); (4) Span selection (in gold). The two encoders here share the same parameters and the two decoders here share the same parameters. The prompts here are simplified and shorten due to the space limit.}
	\label{fig:method}
\end{figure*}

\section{Related Work}

\subsection{Event Argument Extraction}
As a crucial subtask of EE, EAE has long been studied. In the early stages, EAE is only treated as a component of EE systems \citep{chen-etal-2015-event, nguyen-etal-2016-joint-event, yang-etal-2018-dcfee, zheng-etal-2019-doc2edag,lin-etal-2020-joint}, where the phenomenon of event co-occurrence is always taken into account. 

Recently, more and more works study EAE as a stand-alone problem. We summarize these recent works on EAE into 4 categories: (1) span-based methods that identify candidate spans and predict the roles of them \citep{ebner-etal-2020-multi, zhang-etal-2020-two, xu-etal-2022-two}; (2) QA-based methods that query arguments using questions constructed with predefined templates \citep{du-cardie-2020-event, wei-etal-2021-trigger, liu-etal-2021-machine}; (3) sequence-to-sequence methods that leveraging generative PLMs, e.g. BART \citep{lewis-etal-2020-bart} and T5 \citep{JMLR:v21:20-074}, to sequentially generate all arguments of the target event \citep{li-etal-2021-document, du-etal-2021-template, lu-etal-2021-text2event}; (4) a prompt-based method by \citet{ma-etal-2022-prompt} that leverages slotted prompts to extract arguments in a generative slot-filling manner.

Among them, the prompt-based method, PAIE \citep{ma-etal-2022-prompt}, demonstrates SOTA performance. However, all of them only consider one event at a time, diverging from EE research. In this work, we adapt PAIE into a non-autoregressive table generation framework, which is able to extract the arguments of multiple events in parallel.

\subsection{Text-to-Table}
Although table-to-text \cite{10.5555/3504035.3504650, chen-etal-2020-logical} is a well-studied problem in the area of controllable natural language generation, Text-to-Table, the inverse problem of table-to-text, is just newly introduced by \citet{wu-etal-2022-text-table}. In \citet{wu-etal-2022-text-table}, text-to-table is solved with a sequence-to-sequence model enhanced with table constraint and table relation embeddings. In contrast, our table generation framework constructs the slotted table input based on given trigger(s) and predefined prompt(s), and generate in a non-autoregressive manner.

\section{Methodology}
In this section, we will first give an formal definition of EAE and then introduce TabEAE, our solution to the task in detail.

\subsection{Task Definition}
An instance of EAE has the general form of $(\mathbf{x}, \left\{t_i\right\}_{i=1}^{N}, \left\{e_i\right\}_{i=1}^{N}, \left\{R^{e_i}\right\}_{i=1}^{N}, \left\{A_i\right\}_{i=1}^{N})$, where $\mathbf{x}$ is the text (a sentence or a document), $N$ is the number of target events, $t_i$ is the trigger of $i$-th event, $e_i$ is the type of $i$-th event, $R^{e_i}$ is the set of argument roles associated with the event type $e_i$, $A_i$ is the set of arguments of the $i$-th event and each $a^{(r)} \in A_i$ is a textual span within $\mathbf{x}$ that represents the role $r \in R^{e_i}$. Different from the formulation by previous research on EAE that only considers one event for an input instance, this formulation takes all events co-occurring in the same context into consideration, providing a more comprehensive view of the problem.

\subsection{TabEAE}
Our solution to EAE is a non-autoregressive table generation framework, namely TabEAE, which is derived from the SOTA prompt-based EAE model PAIE. Figure \ref{fig:method} gives an overview of the framework. A detailed description of each component comes as follows.

\subsubsection{Trigger-aware Context Encoding}
Given an input text $\mathbf{x} = x_1, x_2, ... , x_L$ with a set of event triggers, we first mark each trigger with a pair of markers $(\text{<T-$i$>}, \text{</T-$i$>})$, where $i$ counts the order of occurrence. Note that, there may be multiple events sharing the same trigger, in which case the shared trigger will only be marked once. After that, we tokenize the marked text into
\begin{align}
    \tilde{\mathbf{x}} = [&\text{<s>}, x_1, x_2, ..., \text{<T-$1$>}, x_{t_1}, \text{</T-$1$>}, \\
    & ..., \text{<T-$i$>}, x_{t_i}, \text{</T-$i$>}, ..., x_L, \text{</s>}]
\end{align}
where $x_{t_i}$ is the text fragment of the $i$-th trigger.

By feeding $\tilde{\mathbf{x}}$ into a transformer-based encoder, we can get the encoding of the text:
\begin{align}
    E_{\tilde{\mathbf{x}}} = \text{Encoder}(\tilde{\mathbf{x}})
\end{align}

We follow PAIE \citep{ma-etal-2022-prompt}, to further decodes $E_{\tilde{\mathbf{x}}}$ with a decoder to obtain the event-oriented context representation:
\begin{align}
    H_{\tilde{\mathbf{x}}} = \text{Decoder}(E_{\tilde{\mathbf{x}}})
    \label{eq4}
\end{align}

\subsubsection{Slotted Table Construction}
The decoder input is constructed as a slotted table, where the column header is the concatenation of event-schema prompt(s) proposed by PAIE. Considering the example in Figure \ref{fig:method}, there are a Life-die event with trigger ``kills'' and a Life-injure event with trigger ``injured''. Then the column header is \emph{``\underline{Victim} (and \underline{Victim}) died at \underline{Place} (and \underline{Place}) killed by \underline{Killer} (and \underline{Killer}). \underline{Victim} (and \underline{Victim}) injured by \underline{Injurer} (and \underline{Injurer}).''}, where the first sentence is the prompt for Life-die event, the second sentence is the prompt for Life-injure event, and each underlined part is named after a argument role, acting as the head of a column. There are multiple columns sharing the same argument role for the extraction of multiple arguments playing the same role in an event.

We initialize the representation of column header by feeding each prompt into the encoder in parallel and concatenating the encoding outputs:
\begin{align}
    E_{\text{PR}_j} &= \text{Encoder}(\text{PR}_j) \\
    E_{\text{CH}} &= [E_{\text{PR}_1}: ...: E_{\text{PR}_j}: ...: E_{\text{PR}_M}]
\end{align}
where $\text{PR}_j$ is the $j$-th prompt, $M$ is the number of event type(s).

The $i$-th row of the table starts with the $i$-th trigger, followed by a sequence of argument slots $S_i$. The initial representation of the $i$-th trigger, $E_{t_i}$, is copied from the encoding of the marked text. And the initial representation of argument slots, $E_{S_i}$, is the average of the encoding of corresponding argument roles (in the column header) and the encoding of corresponding trigger markers. We denote all the argument slots as $\mathbb{S} = \left\{S_i\right\}_{i=1}^N$.

The initial representations of table components are row-wise concatenated to obtain the initial representation of the table:
\begin{align}
    E_{\text{Tab}} &= [E_{\text{CH}} : E_{t_1}\!:\!E_{S_1}\!: ... :\!E_{t_N}\!:\!E_{S_N}\!]
\end{align}

\subsubsection{Non-autoregressive Table Decoding}
The non-autoregressive decoder iteratively updates the representation of input table via structure-aware self-attention inner the table as well as cross-attention between the table and the encoder output.
\paragraph{Structure-aware Self-attention} We devise a structure-aware self-attention mask, $M^{\text{Tab}}$, so that each element of the table can only attend to the region related to it. Our design is as follows:
\begin{itemize}[leftmargin=*]
    \setlength{\itemsep}{1pt}
    \setlength{\parsep}{1pt}
    \setlength{\parskip}{1pt}
    \item All tokens within the column header attend to each other.
    \item All tokens within the column header attend to the event trigger(s).
    \item Each role along with corresponding argument slot(s) attend to each other.
    \item Each event trigger along with corresponding argument slot(s) attend to each other.
\end{itemize}

Note that this attention mask is only used for the decoding of slotted table. When computing $H_{\tilde{\mathbf{x}}}$ (Equation \ref{eq4}), we employ normal self-attention. 

The cross-attention mechanism is the same as the one in Transformer \citep{10.5555/3295222.3295349} and it is only employed to decode the table. When computing $H_{\tilde{\mathbf{x}}}$ (Equation \ref{eq4}), it is skipped.

\subsubsection{Span Selection}
With the output of table decoding, $H_{Tab}$, we can obtain the final representation of the argument slots, $H_\mathbb{S} \subset H_{Tab}$. We transform each representation vector $\mathbf{h}_{s_k} \in H_\mathbb{S}$ into a span selector $\{\Phi_{s_k}^\text{start}, \Phi_{s_k}^\text{end}\}$ \citep{du-cardie-2020-event, ma-etal-2022-prompt}:
\begin{align}
    \Phi_{s_k}^\text{start} &= \mathbf{h}_{s_k} \odot \mathbf{w}^\text{start} \\
    \Phi_{s_k}^\text{end} &= \mathbf{h}_{s_k} \odot \mathbf{w}^\text{end}
\end{align}
where $\mathbf{w}^\text{start}$ and $\mathbf{w}^\text{end}$ are learnable weights, and $\odot$ represents element-wise multiplication.

The span selector $\{\Phi_{s_k}^\text{start}, \Phi_{s_k}^\text{end}\}$ is responsible for selecting a span $(\hat{start}_{k}, \hat{end}_{k})$ from the text to fill in the argument slot $s_k$:
\begin{align}
    \text{logit}_{k}^\text{start} &=  H_{\tilde{\mathbf{x}}} \Phi_{s_k}^\text{start} \in \mathbb{R}^{L} \\
    \text{logit}_{k}^\text{end} &= H_{\tilde{\mathbf{x}}} \Phi_{s_k}^\text{end} \in \mathbb{R}^{L} \\
    \text{score}_{k}(l, m) &= \text{logit}_{k}^\text{start}[l] + \text{logit}_{k}^\text{end}[m] \\
    (\hat{start}_{k}, \hat{end}_{k}) &=  \mathop{\arg\max}\limits_{(l, m):\ 0<m-l<L} \text{score}_{k}(l, m) \nonumber
\end{align}
where $l$ or $m$ represents the index of arbitrary token within the text.

Note that, there can be more than one argument playing the same role in an event, requiring further consideration for the assignment of golden argument spans during training. Hence, we follow \cite{10.1007/978-3-030-58452-8_13, yang-etal-2021-document,ma-etal-2022-prompt} to fine tune our model with the Bipartite Matching Loss. The loss for an training instance is defined as
\begin{align}
    &P_{k}^\text{start} = \text{Softmax} (\text{logit}_{k}^\text{start}) \\
    &P_{k}^\text{end} = \text{Softmax} (\text{logit}_{k}^\text{end}) \\
    \mathcal{L} = - &\sum_{i=1}^N \sum_{(start_{k}, end_{k})\in \delta(A_{i})} (\log P_{k}^\text{start}[start_{k}] \nonumber \\
    & + \log P_{k}^\text{end}[end_{k}])
\end{align}
where $\delta(\cdot)$ represents the optimal assignment calculated with Hungarian algorithm \citep{https://doi.org/10.1002/nav.3800020109} according to the assignment cost devised by \cite{ma-etal-2022-prompt}, and $(start_{k}, end_{k})$ is the golden span optimally assigned to the $k$-th argument slot.  For an argument slot relating to no argument, it is assigned with the empty span $(0, 0)$.

\begin{table*}[htb]
    \centering
    \scalebox{0.87}[0.9]{
    \begin{tabular}{l|l|l|cccccccc}
        \toprule \multirow{2}{*}{Scheme} & \multirow{2}{*}{Model} & \multirow{2}{*}{PLM} & \multicolumn{2}{c}{ACE05} & \multicolumn{2}{c}{RAMS} & \multicolumn{2}{c}{WikiEvents} & \multicolumn{2}{c}{MLEE}  \\
        \cline{4-5} \cline{6-7} \cline{8-9} \cline{10-11} & & & {Arg-I} &  {Arg-C} & {Arg-I} &  {Arg-C} & {Arg-I} &  {Arg-C} & {Arg-I} &  {Arg-C} \\
        \midrule 
        \multirow{7}{*}{Single-Single}
        & {EEQA \shortcite{du-cardie-2020-event}} & BERT & 70.5 & 68.9 & 48.7 & 46.7 & 56.9 & 54.5 & 68.4 & 66.7 \\
        & {EEQA \shortcite{du-cardie-2020-event}}$^\star$ & RoBERTa & 72.1 & 70.4 & 51.9 & 47.5 & 60.4 & 57.2 & 70.3 & 68.7  \\
        & BART-Gen \shortcite{li-etal-2021-document} & BART & 69.9 & 66.7 & 51.2 & 47.1 & 66.8 & 62.4 & 71.0 & 69.8 \\
        & {TSAR \shortcite{xu-etal-2022-two}} & BERT & - & - & 56.1 & 51.2 & 70.8 & 65.5 & 72.3 & 71.3 \\
        & {TSAR \shortcite{xu-etal-2022-two}}$^\star$ & RoBERTa & - & - & 57.0 & 52.1 & \underline{71.1} & 65.8 & \underline{72.6} & \underline{71.5}  \\
        & {PAIE \shortcite{ma-etal-2022-prompt}} & BART & 75.7 & 72.7 & 56.8 & 52.2 & 70.5 & 65.3 & 72.1 & 70.8 \\
        & {PAIE \shortcite{ma-etal-2022-prompt}}$^\star$ & RoBERTa & 76.1 & 73.0 & \underline{57.1} & 52.3 & 70.9 & 65.5 & 72.5 & 71.4 \\
        & {DEGREE \shortcite{hsu-etal-2022-degree}} & BART & 76.0 & 73.5 & - & - & - & - & - & -\\
        & {DEGREE \shortcite{hsu-etal-2022-degree}}$^\star$ & RoBERTa & \underline{76.6} & \underline{73.9} & - & - & - & - & - & - \\
        & TabEAE (Ours) & RoBERTa & 75.5 & 72.6 & 57.0 & \underline{52.5} & 70.8 & 65.4 & 71.9 & 71.0 \\
        \midrule
        \multirow{1}{*}{Multi-Multi} & TabEAE (Ours) & RoBERTa & 75.9 & 73.4 & 56.7 & 51.8 & \underline{71.1} & \underline{66.0} & \textbf{75.1} & \textbf{74.2} \\
        \midrule
        \multirow{1}{*}{Multi-Single} & TabEAE (Ours) & RoBERTa & \textbf{77.2} & \textbf{75.0} & \textbf{57.3} & \textbf{52.7} & \textbf{71.4} & \textbf{66.5} & 72.0 & 71.3 \\
        \bottomrule
    \end{tabular}
    }
    \caption{Main results on four benchmarks. Both RoBERTa and BART here are of large-scale (with 24 Transformer layers). $^\star$ means we replace the original PLM with RoBERTa and rerun their code (hyperparameter tuning is conducted when necessary). The highest scores are in bold font and the second-highest scores are underlined.}
    \label{main_result}
\end{table*}

\subsection{Three Training-Inference Schemes}
Under the TabEAE framework, there exist three possible training-inference schemes: (1) \emph{Single-Single}, train TabEAE to extract single event at a time and infer in the same way; (2) \emph{Multi-Multi}, train TabEAE to extract all events in parallel and infer in the same way; (3) \emph{Multi-Single}, train TabEAE to extract all events in parallel and let it extract single event at a time during inference. For the \emph{Single} mode, only one trigger is marked in the input text; for the \emph{Multi} mode, all the triggers are marked in the text. Note that, when trained to extract all events in parallel, TabEAE also learn to extract single event, since a great portion of training instances has only one event.

\section{Experiments}

\subsection{Implementation Details}
We implement TabEAE with Pytorch and run the experiments with a Nvidia Tesla A100 GPU. We instantiate the encoder with the first 17 layers of RoBERTa-large \citep{DBLP:journals/corr/abs-1907-11692}.\footnote{We choose RoBERTa-large for a fair comparison with EAE methods based on BART-large, as the two PLMs adopt the same tokenizer and are pre-trained on the same corpus.} The weight of the self-attention layers and feedforward layers of the decoder is initialized with the weight of the remaining 7 layers of RoBERTa-large. The setting of 17-layer encoder + 7-layer decoder is found to be optimal by our experiment (See Appendix \ref{appendix:c}). Note that the cross-attention part of the decoder is newly initialized in random and we set its learning rate to be 1.5 times the learning rate of other parameters. We leverage the AdamW optimizer \citep{loshchilov2017decoupled} equipped with a linear learning rate scheduler to tune our model. See Appendix \ref{appendix:b} for details of hyperparameter tuning.

\subsection{Experiment Setups}
\paragraph{Datasets}
We experiment with 4 datasets, including ACE05 \cite{doddington2004the}, RAMS \cite{ebner-etal-2020-multi}, WikiEvents \cite{li-etal-2021-document} and MLEE \cite{10.1093/bioinformatics/bts407}. ACE05 is a sentence-level dataset, while the others are in document-level. The corpora of ACE05, RAMS and WikiEvents mainly consist of news, while the corpus of MLEE lies in the biomedical domain. Besides, the phenomenon of nested event is commonly observed in MLEE, but rare in the other 3 datasets. See Appendix \ref{appendix:a} for a detailed description of the datasets.

\paragraph{Evaluation Metrics}
Following previous works \citep{li-etal-2021-document, ma-etal-2022-prompt}, we measure the performance with two metrics: (1) strict argument identification F1 (Arg-I), where a predicted argument of an event is correct if its boundary matches any golden arguments of the event; (2) strict argument classification F1 (Arg-C), where a predicted argument of an event is correct only if its boundary and role type are both correct. All the reported results are averaged over 5 runs with different random seeds.

\subsection{Compared Methods}
We compare TabEAE with several SOTA methods:
\begin{itemize}[itemsep=2pt,topsep=1pt,parsep=0pt]
    \item \textbf{EEQA} \cite{du-cardie-2020-event}, a QA-based EAE model that treats EAE as a machine reading comprehension problem;
    \item \textbf{BART-Gen} \cite{li-etal-2021-document}, a seq-to-seq EAE model that generates predicted arguments conditioned on event template and context;
    \item \textbf{TSAR} \cite{xu-etal-2022-two}, a two-stream AMR-enhanced span-based EAE model;
    \item \textbf{PAIE} \cite{ma-etal-2022-prompt}, a prompt-based EAE model that leverages slotted prompts to obtain argument span selectors;
    \item \textbf{DEGREE} \cite{hsu-etal-2022-degree}, a data-efficient model that formulates EAE as a conditional generation problem.
\end{itemize}

\begin{table*}[htb]
    \centering
    \scalebox{0.85}[0.87]{
    \begin{tabular}{l|l|cccccccc}
        \toprule \multirow{3}{*}{Model} & \multirow{3}{*}{Scheme} & \multicolumn{2}{c}{ACE05} & \multicolumn{2}{c}{RAMS} & \multicolumn{2}{c}{WikiEvents} & \multicolumn{2}{c}{MLEE}  \\
        \cline{3-4} \cline{5-6} \cline{7-8} \cline{9-10} & & {\# Ev = 1} &  {\# Ev > 1} & {\# Ev = 1} &  {\# Ev > 1} & {\# Ev = 1} &  {\# Ev > 1} & {\# Ev = 1} &  {\# Ev > 1} \\
        & & [185] & [218] & [587] & [284] & [114] & [251] & [175] & [2025] \\
        \midrule
        PAIE \shortcite{ma-etal-2022-prompt} & Single-Single & 70.97 & 73.88 & 52.72 & 52.14 & 65.31 & 65.37 & 78.91 & 70.11  \\
        \midrule
        \multirow{3}{*}{TabEAE} & Single-Single & 71.21 & 73.83 & 52.82 & 51.61 & 65.27 & 65.46 & 79.26 & 70.32 \\
        \cline{2-10}
        & Multi-Multi & \multirow{2}{*}{\textbf{73.38}} & 73.45 & \multirow{2}{*}{\textbf{52.87}} & 50.82 & \multirow{2}{*}{\textbf{67.30}} & 65.32 & \multirow{2}{*}{\textbf{81.13}} & \textbf{73.60} \\
        & Multi-Single & & \textbf{76.13} & & \textbf{52.49} & & \textbf{66.19} & & 69.97 \\
        \bottomrule
    \end{tabular}
    }
    \caption{Comparison of EAE models with different training-inference schemes in terms of their performance on instances with different numbers of events. The number in brackets is the number of supporting events in the test set. Unless otherwise specified, we only measure the Arg-C F1 for the experiments in the Analysis section.}
    \label{diff_num_event}
    \vspace{-0.5em}
\end{table*}

\begin{figure}[!h]
\centering

\subfigure[ACE05] {
\includegraphics[height = 0.465 \columnwidth,width=0.465\columnwidth]{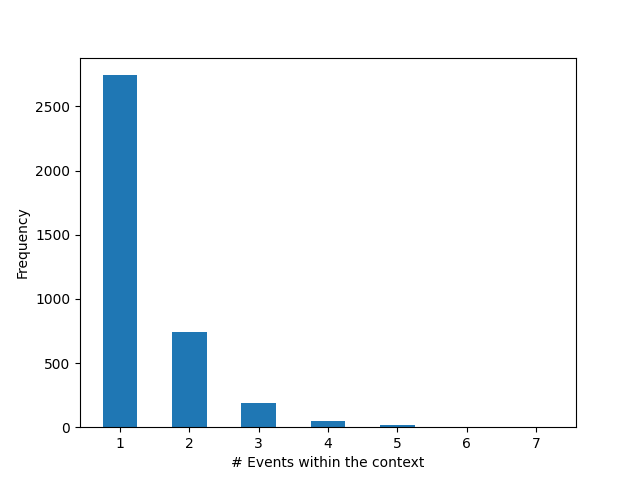}
}\vspace{-3.5mm}
\subfigure[RAMS] {
\includegraphics[height = 0.465 \columnwidth, width=0.465\columnwidth]{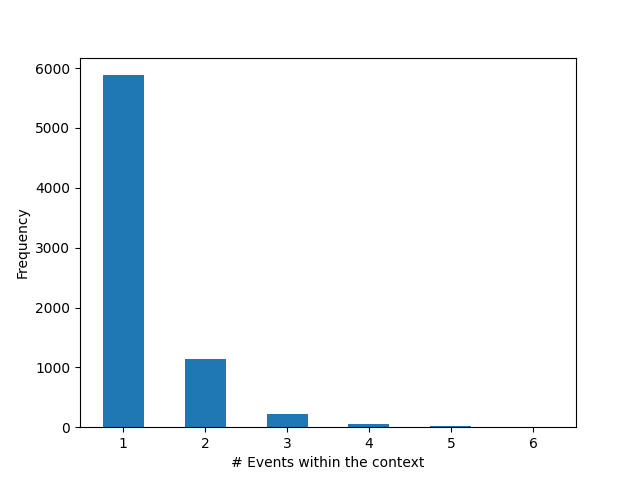}
} 
\subfigure[WikiEvents] {
\includegraphics[height = 0.465 \columnwidth,width=0.465\columnwidth]{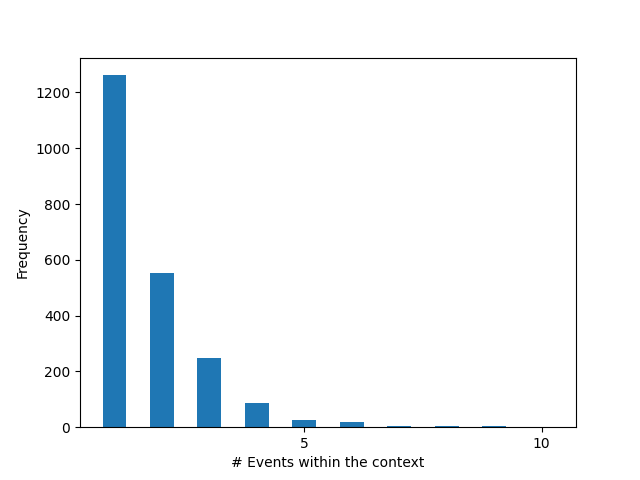}
}\vspace{-1.5mm}
\subfigure[MLEE] {
\includegraphics[height = 0.465 \columnwidth,width=0.465\columnwidth]{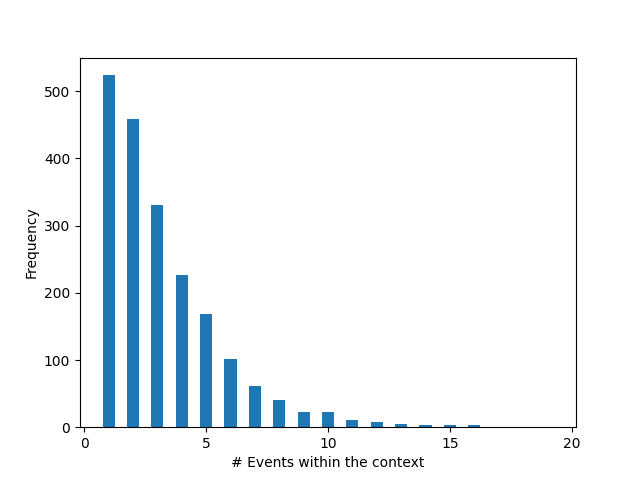}
}
\caption{Distributions of the number of events per instance on the four datasets.}
\label{freq}
\vspace{-0.5em}
\end{figure}

\subsection{Main Results}
The overall performances of compared baselines and TabEAE are illustrated in Table \ref{main_result}. We find that TabEAE (Single-Single) is competitive to previous SOTA models (TSAR, DEGREE and PAIE) on the four benchmarks. This is expected since these models follow the same training-inference scheme and leverage PLMs of the same scale.

In the mean time, TabEAE (Multi-Single) outperforms the SOTA model by 0.6 Arg-I F1 and 1.1 Arg-C F1 on ACE05, by 0.2 Arg-I F1 and 0.4 Arg-C F1 on RAMS, by 0.3 Arg-I F1 and 0.7 Arg-C F1 WikiEvents.

As for the MLEE dataset, TabEAE (Multi-Multi) performs better than TabEAE (Multi-Single) and yields 2.5 Arg-I F1 gain, 2.7 Arg-C F1 gain compared to SOTA models.

We analyze the reason behind the results in \S \ref{Training-inference_Scheme}.

\begin{table*}[htb]
    \centering
    \scalebox{0.825}[0.85]{
    \begin{tabular}{l|cccccccc}
        \toprule \multirow{3}{*}{Scheme} & \multicolumn{2}{c}{ACE05} & \multicolumn{2}{c}{RAMS} & \multicolumn{2}{c}{WikiEvents} & \multicolumn{2}{c}{MLEE}  \\
        \cline{2-3} \cline{4-5} \cline{6-7} \cline{8-9} & {N-O} & {Overlap} & {N-O} & {Overlap} & {N-O} & {Overlap} & {N-O} & {Overlap} \\
        & [319] & [84] & [690] & [181] & [296] & [69] & [1460] &  [734] \\
        \midrule 
        Single-Single & 71.1 & 78.6 & 51.6 & 55.6 & 65.7 & 64.4 & 75.4 & 65.8 \\
        Multi-Multi & - & - & - & - & - & - & 78.1 (+2.7) & 69.4 (+3.6) \\
        Multi-Single & 72.8 (+1.7) & 80.8 (+2.2) & 51.7 (+0.1) & 56.1 (+0.5) & 66.4 (+0.7) & 66.9 (+2.5) & - & - \\
        \bottomrule
    \end{tabular}
    }
    \caption{Comparison of TabEAE with different training-inference schemes in terms of their  capacity to extract the arguments of overlapping events (events with shared arguments). The number in square brackets is the number of supporting events in the test set.  \textbf{N-O}: Non-overlapping.}
    \label{Overlapping}
    \vspace{-0.5em}
\end{table*}

\section{Analysis}
\subsection{The Effect of Training-inference Scheme}
\label{Training-inference_Scheme}
To analyze the influence of the training-inference scheme, we measure the performances of EAE models with different training-inference schemes on handling instances with different numbers of events. The results are shown in Table \ref{diff_num_event}. We can see that PAIE (Single-Single) and TabEAE (Single-single) have similar capacity in extracting stand-alone events and co-occurring events.

When trained to extract all the events in parallel, the Arg-C F1 of TabEAE on instances with single event increases by 2.17, 0.05, 2.03 and 1.87 on the 4 datasets respectively. However, by letting TabEAE extract all the events in parallel during inference, the Arg-C F1 on instances with multiple events drops by 0.38, 0.79, 0.14 on ACE, RAMS and WikiEvents respectively, while increasing by 3.28 on MLEE. We believe this phenomenon is the result of two factors:
\begin{enumerate}[itemsep=2pt,topsep=1pt,parsep=0pt]
    \item The distribution of the number of events per instance. As plotted in Figure \ref{freq}, there are much more instances with multiple events on WikiEvents and MLEE than on ACE05 and RAMS. Hence, the model is better trained to extract multiple events concurrently on WikiEvents and MLEE.

    \item Difficulty. Generally, it is more difficult for a model to extract all the events in one pass. But it is not the case for the MLEE dataset, since there are around 32.9\% of the arguments acting as triggers of other events in MLEE, and when all triggers are provided (as in the Multi-Multi scheme), it become easier for the model to extract all the arguments.
\end{enumerate}

When training TabEAE to extract all events in parallel and letting it extract one event at a time during inference, the Arg-C F1 of TabEAE on instances with multiple events increases by 2.3, 0.88, 0.73 on ACE, RAMS and WikiEvents respectively. This is reasonable, since there is a large portion of instances having only one event, which means the model is also well-trained to extract one event at a time under the Multi-Single scheme.

\subsection{Capturing the Event Semantic Boundary}

We hypothesize that the performance gains yielded by the Multi-Multi and Multi-Single schemes rooted in the stronger ability of TabEAE to capture the event semantic boundary. To verify this, we further measure the model's ability to capture the event semantic boundary from two points of view: (1) Inter-event Semantic; (2) Inner-event Semantic.

From the view of \textbf{inter-event semantic}, we compare the performance of TabEAE with different training-inference schemes in terms of their ability to extract the arguments of overlapping events (i.e., events with shared arguments). As illustrated in Table \ref{Overlapping}, when trained to extract all events concurrently, the model's performance gains of extracting the arguments of overlapping events are much higher than that of extracting the arguments of non-overlapping events. Specifically, the differences of performance gains are 0.5 Arg-C F1 on ACE05, 0.4 Arg-C F1 on RMAS, 1.8 Arg-C F1 on WikiEvents and 0.9 Arg-C F1 on MLEE. This suggests that TabEAE can better distinguish the semantic boundary between overlapping events.

\begin{figure}[ht]
\centering

\subfigure[WikiEvents] {
\includegraphics[width=0.465\linewidth]{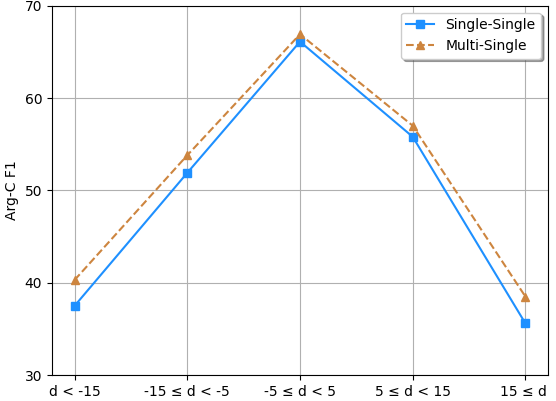}
}
\subfigure[MLEE] {
\includegraphics[width=0.465\linewidth]{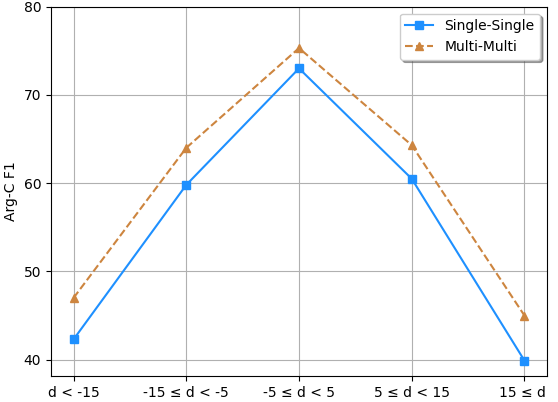}
}

\caption{Comparison of TabEAE with different training-inference schemes in terms of their ability to extract arguments of different distances to the triggers. $d$ is the word distance from a trigger to an argument (the value is negative when the argument is on the left side of the trigger).}
\label{distance}
\end{figure}

From the view of \textbf{inner-event semantic}, we compare the performance of TabEAE with different training-inference schemes in terms of their ability to extract arguments of different distances to the triggers. We define the distance here as the head word index of an argument minus the head word index of its corresponding trigger. The experiments are conducted on the document-level datasets WikiEvents and MLEE, where the distance distribution of event arguments is more disperse. The results are plotted in Figure \ref{distance}. We can observe that, when equipped with the Multi-Multi/Multi-Single schemes the model's performance gains of extracting remote arguments are higher than the performance gains of extracting nearby arguments. This means TabEAE gets better at extracting arguments around the event boundary.

\begin{table}[ht]
    \centering
    \scalebox{0.80}[0.84]{
        \begin{tabular}{l|cccc}
        \toprule {Model} & {ACE05} & {RAMS} & {WikiEvents} & {MLEE} \\
        \midrule
        TabEAE & 75.0 & 52.7 & 66.5 & 74.2 \\
        \midrule 
        \ w/o SAAM & 73.1 & 51.2 & 65.4 & 72.7\\
        \ w/o PET & 70.8 & 49.3 & 61.9 & 69.9\\
        \ w/o Prompts & 72.5 & 50.9 & 64.8 & 71.3 \\
        \ BERT $\rightarrow$ BART & 72.7 & 51.0 & 65.4 & 72.4 \\
        \bottomrule
    \end{tabular}
    }
    \caption{Results of ablation study. For ACE05, RAMS and WikiEvents, we experiment with TabEAE (Multi-Single); for MLEE, we experiment with TabEAE (Multi-Multi). SAAM: Structure-aware Attention Mask. PET: Pre-computed Encodings of the input Table.}
    \label{ablation_study}
    \vspace{-0.5em}
\end{table}

\begin{figure}[h!]
    \centering
    \includegraphics[width=0.9\columnwidth]{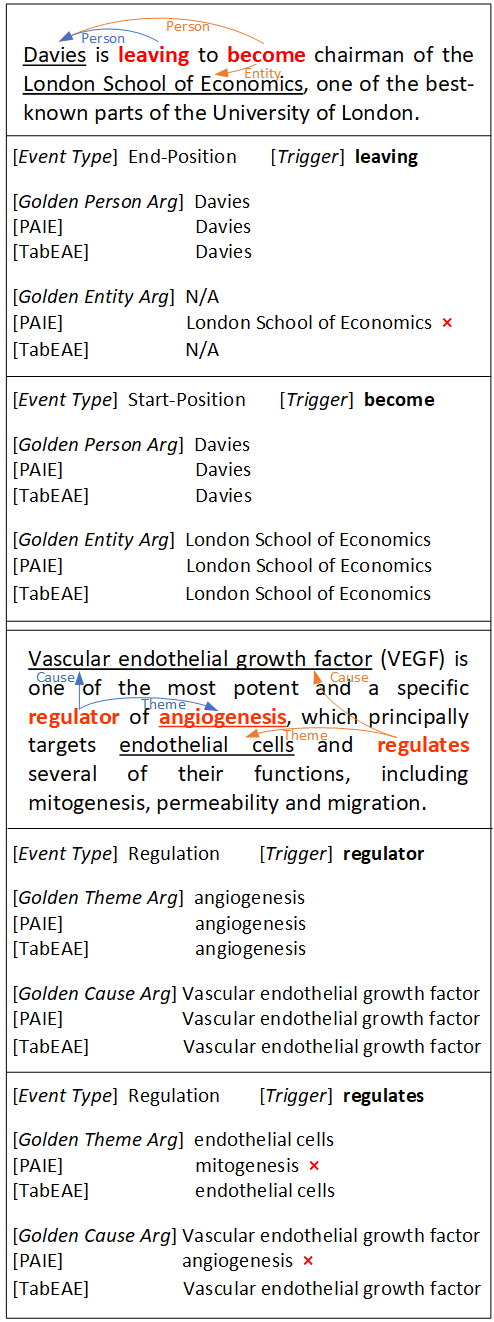}
    \caption{Two test cases from ACE05 and MLEE.}
    \label{fig:case}
    \vspace{-0.5em}
\end{figure}

\subsection{Ablation Study}
To verify the effectiveness of different components of TabEAE, we conduct ablation study on the 4 datasets. The results are illustrated in Table \ref{ablation_study}.

After removing the \textbf{structure-aware attention mask}, the Arg-C F1 scores drop by 1.9, 1.5, 1.1, 1.5 on ACE05, RAMS, WikiEvents and MLEE respectively. This demonstrates the benefit of letting each table token only paying attention to the table region related to it.

After replacing the \textbf{pre-computed encodings of the input table} with RoBERTa token embeddings, the Arg-C F1 scores drop by 4.2, 3.4, 4.6, 3.9 on the 4 datasets. This proves the necessity of initializing the embeddings of input table with the encodings computed by the encoder.

When constructing the table column header with the concatenation of argument roles instead of \textbf{prompts}, the Arg-C F1 scores drop by 2.5, 1.8, 1.7 and 2.9 on the 4 datasets respectively. This coincides with the finding by \cite{ma-etal-2022-prompt} that hand-crafted prompts can be of great help to the task of EAE.

When replacing the encoder/decoder of TabEAE with \textbf{BART} encoder/decoder, the model performance degrades by 2.3, 1.7, 1.1, 1.8 on the 4 datasets respectively. The reason behind this degradation should be the uni-directional self-attention employed by BART decoder is not suitable for the decoding of table.

\subsection{Case Study}
Figure \ref{fig:case} illustrates 2 test cases from ACE05 and MLEE respectively. In the first test case, there exist 2 events triggered by ``leaving'' and ``become'', with a shared argument ``Davies''. PAIE incorrectly predicts ``London School of Economics'' as an argument of the event triggered by ``leaving'', which is essentially an argument of the event triggered by ``become''. In contrast, TabEAE is able to avoid this mistake, demonstrating a stronger capacity to capture the event semantic boundary.

In the second test case, there exist 3 events triggered by ``regulator'', ``regulates'' and ``angiogenesis'' respectively. Among them, the event triggered by ``angiogenesis'' has no argument. For the event triggered by ``regulates'', PAIE fails to extract the remote argument ``Vascular endothelial growth factor'', while TabEAE correctly extracts it by being aware of the co-occurring event that shares this argument.

\section{Conclusion}
In this paper, we point out that recent studies on EAE ignore event co-occurrences, resulting in a divergence from main-stream EE research. To remedy this, we highlight the question that \emph{``Can EAE models learn better when being aware of event co-occurrences''} and explore it with a novel text-to-table framework, \emph{TabEAE}, that can extract multiple event in parallel. By experimenting with 3 training-inference schemes on 4 datasets, we find that when trained to extract all event concurrently, TabEAE can better capture the event semantic boundary and its ability to extract single event gets greatly improved. Our work demonstrates the significance of event co-occurrence for EAE and establishes a new foundation for future EAE research.

\section{Limitations}
In this section, we summarize the limitations of our work as follows:
\begin{itemize}[leftmargin=*]
    \setlength{\itemsep}{2pt}
    \setlength{\parsep}{2pt}
    \setlength{\parskip}{2pt}
    \item There is still a lot more to explore in terms of event co-occurrence for EAE (e.g., iterative extraction, course learning, etc.). We are unable to cover all in this work and will explore further in the future.
    \item As demonstrated by our ablation study, the high performance of our model greatly relies on the manual prompts. This limits the application of our model to the scenes where high-quality prompts are unavailable and difficult to construct. To address this, we should look into the area of automatic prompt construction.
    \item Our work ignores the phenomenon of entity co-reference commonly existing in narrative documents. This limits the model's ability to figure out the underlying relation between entities, which is crucial for the task of EAE. And we will take entity co-references into account in our future works.
\end{itemize}

\section*{Acknowledgments}
We thank the reviewers for their insightful comments and valuable suggestions. This study is partially supported by National Key R\&D Program of China (2021ZD0113402), National Natural Science Foundations of China (62276082, U1813215 and 61876052), National Natural Science Foundation of Guangdong, China (2019A1515011158), Major Key Project of PCL (PCL2021A06), Strategic Emerging Industry Development Special Fund of Shenzhen (20200821174109001) and Pilot Project in 5G + Health Application of Ministry of Industry and Information Technology \& National Health Commission (5G + Luohu Hospital Group: an Attempt to New Health Management Styles of Residents).

\bibliography{main}

\begin{table*}[h!]
	\centering
	\scalebox{1.0}[1.0]{
	\begin{tabular}{l|cccc}
		\toprule \textbf{Dataset} & \textbf{ACE05} & \textbf{RAMS} & \textbf{WikiEvents} & \textbf{MLEE}  \\
		\midrule
		\textbf{\# Event types} & 33 & 139 & 50 & 23 \\
		\textbf{\# Args per event} & 1.19 & 2.33 & 1.40 & 1.29 \\
		\textbf{\# Events per text} & 1.35 & 1.25 & 1.78 & 3.32 \\
		\midrule
		\textbf{\# Events} \\
		Train & 4202 & 7329 & 3241 & 4442\\
		Dev & 450 & 924 & 345 & - \\
		Test & 403 & 871 & 365 & 2200 \\
		\bottomrule
	\end{tabular}
	}
	\caption{Dataset Statistics.}
	\label{statistic}
\end{table*}

\appendix

\section{Profile of Datasets}
\label{appendix:a}
\paragraph{ACE05}\citep{doddington2004the}\footnote{https://catalog.ldc.upenn.edu/LDC2006T06} is an annotated information extraction corpus of newswire, broadcast news and telephone conversations. We utilize its English event annotation for sentence-level EAE. We preprocess the data in the same way as \citet{wadden-etal-2019-entity} do.
\paragraph{RAMS}\citep{ebner-etal-2020-multi}\footnote{https://nlp.jhu.edu/rams/} is a document-level EAE dataset, which contains 9,124 annotated events from English online news. Since it is annotated event-wise (each event occupies one instance), we have to aggregate events occurring in the same context into one instance with multiple events. We follow the original train/dev/test data split.
\paragraph{WikiEvents}\citep{li-etal-2021-document}\footnote{https://github.com/raspberryice/gen-arg} is a document-level EAE dataset, consisting of events recorded in English Wikipedia along with the linking news articles that mention these events. The dataset is also annotated with the co-reference links of arguments, but we only use the exact argument annotations in our experiments.
\paragraph{MLEE}\citep{10.1093/bioinformatics/bts407}\footnote{http://www.nactem.ac.uk/MLEE/} is a document-level event extraction dataset with manually annotated abstracts of bio-medical publications written in English. We follow the preprocessing procedure of \citep{10.1093/bioinformatics/btaa540}. Since there is only train/test data split for the preprocessed dataset, we employ the training set as the development set.

\paragraph{Statistics} Detailed statistics of the datasets are listed in Table \ref{statistic}.

\section{hyperparameter Settings}
\label{appendix:b}
Most of the hyperparameters follow the same configuration of \cite{ma-etal-2022-prompt}. We only tune a few hyperparameters manually for each dataset by trying different values of each hyperparameter within an interval and choosing the value that results in the highest Arg-C F1 on the development set. The trial-intervals and the final hyperparameter configuration are shown in Table \ref{hyperparameter}.

\begin{table*}[ht]
    \centering
    \scalebox{1}[1]{
    \begin{tabular}{ccccccc}
    \toprule \textbf{hyperparameters} & \textbf{Trial-Interval} & \textbf{ACE05} & \textbf{RAMS} & \textbf{WikiEvents} & \textbf{MLEE}\\
    \midrule
    Training Steps & - & 10000 & 10000 & 10000 & 10000 \\
    Warmup Ratio & - & 0.1 & 0.1 & 0.1 & 0.1 \\
    Learning Rate & - & 2e-5 & 2e-5 & 2e-5 & 2e-5 \\
    Max Gradient Norm & - & 5 & 5 & 5 & 5 \\
    Batch Size & [2, 16] & 8 & 4 & 4 & 4 \\
    Context Window Size & - & 250 & 250 & 250 & 250 \\
    Max Span Length & - & 10 & 10 & 10 & 10 \\
    Max Encoder Seq Length & - & 200 & 500 & 500 & 500 \\
    Max Decoder Seq Length & [200, 400] & 250 & 200 & 360 & 360 \\
    \bottomrule
    \end{tabular}
    }
    \caption{hyperparameter settings. The hyperparameters without trial-interval are set to be the same as in \citet{ma-etal-2022-prompt} without tuning.}
    \label{hyperparameter}
\end{table*}

\begin{figure}
    \centering
    \includegraphics[width=0.93\columnwidth]{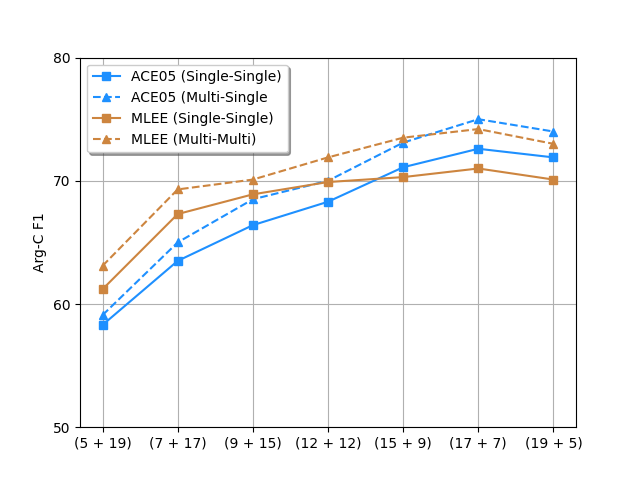}
    \caption{Comparison of TabEAE with different numbers of encoder-decoder layers. $(x + y)$ represents $x$-layer encoder + $y$-layer decoder.}
    \label{num_layers}
\end{figure}

\section{Number of Encoder/Decoder Layers}
\label{appendix:c}
We have employed the bottom layers of RoBERTa-large as our encoder and the top layers of RoBERTa-large as our decoder. To find the optimal layer allocation, we have tried different settings and recorded the corresponding model performance. This experiment is conduct on ACE and MLEE. The results are plotted in Figure \ref{num_layers}. We can observe that the overall performance on the two datasets reaches the peak when there are 17 encoder layers and 7 decoder layers in the model. This observation coincides with recent findings on the areas of machine translation and spell checking that ``deep encoder + shallow decoder'' is superior to the conventional architecture with balanced encoder-decoder depth \citep{Kasai2020DeepES, sun-etal-2021-instantaneous}.

\section{Prompt Construction}
\label{appendix:d}

The prompts for ACE05, RAMS and WikiEvents are directly from \citep{li-etal-2021-document, ma-etal-2022-prompt}, which are manually constructed from the pre-defined ontology associated with each dataset. For MLEE, we manually construct the prompts in a similar manner, as shown in Table \ref{prompt}.

\begin{table*}[h!]
	\centering
	\scalebox{0.9}[0.92]{
	\begin{tabular}{l|l}
		\toprule \textbf{Event Type} & \textbf{Natural Language Prompt}  \\
		\midrule
		Cell proliferation & \underline{Cell} proliferate or accumulate \\
		Development & \underline{Anatomical Entity} develop or form \\
		Blood vessel development & neovascularization or angiogenesis at \underline{Anatomical Location} \\
		Growth & growth of \underline{Anatomical Entity} \\
		Death & death of \underline{Anatomical Entity} \\
		Breakdown & \underline{Anatomical Entity} degraded or damaged \\
		Remodeling & \underline{Tissue} remodeling or changes \\
		Synthesis & synthesis of \underline{Drug/Compound} \\
		Gene expression & expression of \underline{Gene} and \underline{Gene} ( and \underline{Gene} ) \\
        Transcription & transcription of \underline{Gene} \\
        Protein processing & processing of \underline{Gene product} \\
        DNA methylation & methylation of \underline{Entity} at \underline{Site} \\
        Metabolism & metabolism of \underline{Entity} \\
        Catabolism & catabolism of \underline{Entity} \\
        Phosphorylation & phosphorylation of \underline{Entity} at \underline{Site} \\
        Dephosphorylation & dephosphorylation of \underline{Entity} at \underline{Site} \\
        Pathway & \underline{Entity} and \underline{Entity} and \underline{Entity} ( and \underline{Entity} ) participate in signaling pathway or system \\
        Localization & \underline{Entity} \underline{At Location} or \underline{To Location} or \underline{From Location} \\ 
        Binding & \underline{Site} of \underline{Entity} bind or interact with \underline{Site} of \underline{Entity} ( and \underline{Site} of \underline{Entity} ) \\
        Regulation & \underline{Something} regulate \underline{Event/Entity} at \underline{Site} \\
        Positive regulation & \underline{Something} positively regulate \underline{Event/Entity} at \underline{Site} \\
        Negative regulation & \underline{Something} negatively regulate \underline{Event/Entity} at \underline{Site} \\
        Planned process & \underline{Something} is treated with \underline{Entity} and \underline{Entity} ( and \underline{Entity} ) \\
		\bottomrule
	\end{tabular}
	}
	\caption{Prompts manually constructed for the MLEE dataset.}
	\label{prompt}
\end{table*}

\end{document}